\documentclass[journal]{IEEEtran}

\usepackage{cite}

\ifCLASSINFOpdf
  \usepackage[pdftex]{graphicx}
\else
\fi
\usepackage{amsmath, amssymb}
\usepackage{algorithmic,algorithm}
\usepackage{array}
\ifCLASSOPTIONcompsoc
 \usepackage[caption=false,font=normalsize,labelfont=sf,textfont=sf]{subfig}
\else
 \usepackage[caption=false,font=footnotesize]{subfig}
\fi
\usepackage{url}
\usepackage{enumerate}
\usepackage{enumitem}
\usepackage{multirow}
\usepackage{makecell}
\usepackage{booktabs} % for professional tables
\usepackage[table]{xcolor}
\definecolor{Gray}{gray}{0.9}
\usepackage{tikz}
\usetikzlibrary{shadings}

\begin{document}

\title{ESW Edge-Weights : Ensemble Stochastic Watershed Edge-Weights for Hyperspectral Image Classification}

\author{Rohan~Agarwal,
        Aman~Aziz,
        Aditya~Suraj~Krishnan,
        Aditya~Challa,
        Sravan~Danda% <-this % stops a space
\thanks{All the authors are with Computer Science and information Systems, BITS Pilani K K Birla Goa Campus. Aditya Challa and Sravan Danda are also affiliated to APPCAIR centre. Email ids (in order of authors):  agarwalrohan189@gmail.com, f20180119@goa.bits-pilani.ac.in, adityasuraj99@gmail.com, aditya.challa.20@gmail.com, sravan8809@gmail.com}% <-this % stops a space
}
\markboth{}%
{ESW Edge-Weights : Ensemble Stochastic Watershed Edge-Weights for Hyperspectral Image Classification}

\maketitle

\begin{abstract}
Hyperspectral image (HSI) classification is a topic of active research. One of the main challenges of HSI classification is the lack of reliable labelled samples. Various semi-supervised and unsupervised classification methods are proposed to handle the low number of labelled samples. Chief among them are graph convolution networks (GCN) and their variants. These approaches exploit the graph structure for semi-supervised and unsupervised classification. While several of these methods implicitly construct edge-weights, to our knowledge, not much work has been done to estimate the edge-weights explicitly. In this article, we estimate the edge-weights explicitly and use them for the downstream classification tasks - both semi-supervised and unsupervised. The proposed edge-weights are based on two key insights - (a) Ensembles reduce the variance and (b) Classes in HSI datasets and feature similarity have only one-sided implications. That is, while same classes would have similar features, similar features do not necessarily imply same classes. Exploiting these, we estimate the edge-weights using an aggregate of ensembles of watersheds over subsamples of features. These edge weights are evaluated for both semi-supervised and unsupervised classification tasks. The evaluation for semi-supervised tasks uses Random-Walk based approach. For the unsupervised case, we use a simple filter using a graph convolution network (GCN). In both these cases, the proposed edge weights outperform the traditional approaches to compute edge-weights - Euclidean distances and cosine similarities. Fascinatingly, with the proposed edge-weights, the simplest GCN obtained results comparable to the recent state-of-the-art.
\end{abstract}


\IEEEpeerreviewmaketitle

\section{Introduction}
\IEEEPARstart{H}{yperspectral} Image (HSI) Classification is an area of active research \cite{IEEE:jounral/tgrs/Nan21,IEEE:jounral/tgrs/Yaoming21,SPRINGER:journal/Multimedia/Patel22,IEEE:journal/grsm/Ghamisi17}, thanks to it's wide application ranging from mineral exploration \cite{IEEE:journal/tgrs/Kruse03} to military reconnaissance \cite{IEEE:journal/tgrs/Shimoni19}. This is due to the rich spatial and spectral information available within an HSI dataset. However, HSI classification depends largely on the ability to obtain noise-free ground-truth labels. This is usually costly and sometimes infeasible. Hence, several research studies focussed on semi-supervised or unsupervised classification techniques.

{  
Semi-supervised methods try to use the large number of unlabelled data points along with limited labelled data for classification. A common approach is \emph{active learning} where the data points are actively selected and labelled \cite{IEEE:journal/grsl/Lixia14,IEEE:journal/jstars/Chen19,IEEE:journal/tip/Hao18}. This procedure is iteratively repeated until all points are labelled. On the other hand, unsupervised methods do not utilize the labelled data but instead identify the conspicuous classes within the dataset and map them to groundtruth labels. \cite{IEEE:journal/grsm/Han21} provides a detailed review of the clustering approaches used for HSI clustering. 

Graph based methods for HSI classification have been widely used for both semi-supervised or unsupervised approaches. \cite{IEEE:journal/tgrs/Valls07} proposes a semi-supervised graph based HSI classification. \cite{IEEE:journal/tgrs/Yao21} proposes semi-supervised graph neural network, \cite{IEEE:journal/jstars/Haoyu21, ELSIVIER:journal/pr/Sellami22} utilizes graph convolution networks for semi-supervised HSI classification. In case of unsupervised approaches, the most widely used approach is that of spectral clustering \cite{SPRINGER:journal/statcomp/Luxburg07}. Recently in \cite{IEEE:jounral/tgrs/Nan21,IEEE:jounral/tgrs/Yaoming21} graph based methods are combined with other approaches such as subspace clustering to obtain better results. \cite{IEEE:journal/tgrs/Luo16} constructs the graph using manifold-based sparse representation and graph embedding. \cite{IEEE:journal/tgrs/Luo22} proposes multi-structure unified discriminative embedding (MUDE) which designs the intraclass and interclass neighborhood structure graphs.}

\begin{figure}
  \centering
  \subfloat[Indianpines GT]{\includegraphics[width=0.35\linewidth]{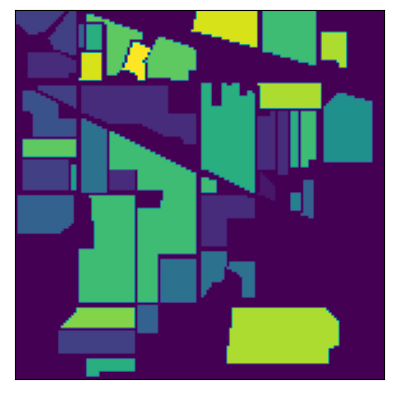}%
  \label{fig:1a}}
  \hfil
  \subfloat[Salinas GT]{\includegraphics[width=0.15\linewidth]{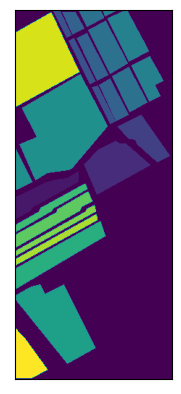}%
  \label{fig:1b}}

  \subfloat[Indianpines GT]{\includegraphics[width=0.35\linewidth]{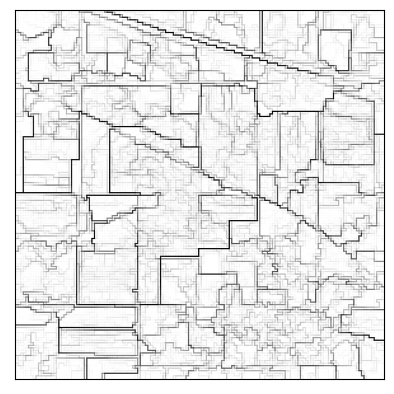}%
  \label{fig:1c}}
  \hfil
  \subfloat[Salinas GT]{\includegraphics[width=0.15\linewidth]{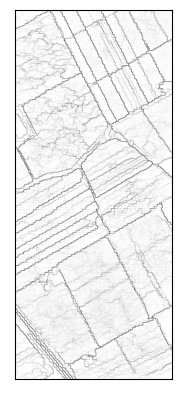}%
  \label{fig:1d}}

  \caption{Visualizing the ESW Edge-Weights. (a) Groundtruth objects in the Indianpines dataset. (b) Groundtruth objects in Salinas dataset. (c) Proposed ESW Edge-Weights on Indianpines dataset. (d) Proposed ESW Edge-weights on Salinas dataset. Observe that the boundaries are easily discernible in the edge weights plots. To visualize the edge-weights we use the 2D cubical complexes\cite{SCIENCEDIRECT:journal/cvgip/kovalevsky89}.}
  \label{fig:1}

\end{figure}

{ While several graph based approaches implicitly compute edge-weights, we are not aware of any works which explicitly estimates the edge-weights.} In this article \emph{we propose an approach to estimate the edge-weights for HSI datasets} referred to as \emph{ESW Edge-Weights} - \emph{Ensemble Stochastic Watersheds based Edge-Weights}. Figure \ref{fig:1} visualizes the ESW Edge-Weights obtained by our proposed approach. Observe that the object boundaries are easily discernible.  We show that ESW edge-weights perform better than existing approaches such as Euclidean distances or cosine similarities for both semi-supervised and unsupervised classification. Moreover, we also show that using the proposed edge-weights, even the simplest GCN approach results in scores better than the recent state-of-the-art.

% As illustrated in Figures 5(e,f), a visualization of a saliency map when the graph is given by the 4- adjacency relation can be obtained thanks to cubical complexes (also known as Khalimsky grids). Cubical complexes have been promoted in particular by V. Kovalevsky [25] in order to provide a sound topological basis for image analysis. In 2D, a cubical complex is a set of squares, unit line segments (represented by rectangles in Figure 5(e)), and unit points (represented by dots Figure 5(e)). Each vertex of the graph can be identified to a square of the complex. Then, each edge linking two vertices x and y can be identified to the segment corresponding to the common side of the two squares identified with x and y. The squares are given a null value whereas the sides are given the value of the associated edges in the saliency map. Finally, for each point of the complex (i.e., the corners of the squares), the maximal value of a side containing it is kept. Thus, any element of the complex has a value. Hence, since the elements of the complex are aligned on a square matrix, the saliency map can be visualized as an image (see Figure 5(f)).

{
The main contributions in this article are as follows:
\begin{enumerate}[label=(\roman*)]
  \item We propose a novel approach to estimate the edge-weights within HSI datasets, referred to as \emph{ESW Edge-Weights}.
  \item We show that these edge-weights are superior than the classical ones such as Euclidean distance or cosine similarity using Random Walk semi-supervised classification.
  \item We show how vanilla graph convolution networks (GCNs) (as proposed in \cite{ACM:conf/ijcai/Xiaotong19}) can be modified to incorporate edge-weights. This results in a unsupervised learning scheme better than the unweighted approach. Moreover we show that using the proposed edge-weights, we obtain results which are better than the recent state-of-the-art.
\end{enumerate}}

\section{Estimate Edge Weights using Ensembles of Stochastic Watersheds}
{\noindent
\textbf{Notation:} Let $X$ denote the HSI dataset with $nc$ columns, $nr$ rows and $nz$ bands. Let $G = (V, E, W)$ denote the edge-weighted graph, where $V$ denotes the set of vertices. Each vertex corresponds to a pixel within HSI dataset. Hence for the graph corresponding to dataset $X$, one would have $nc \times nr$ number of vertices. $E \subseteq V\times V$ denotes the subset of edges, and $W : E \to \mathbb{R}^{+}$ denotes the set of edge weights. In the rest of this article we use the 4-adjacency edges as $E$, unless explicitly stated otherwise.
}

In this section we ask and answer the question - Given and edge $e=(e_x,e_y)$, what is the best estimate of the edge weight $W(e)$? An ideal edge-weight would reflect the chance that the vertices in the edge belong to a different class. That is,
\begin{equation}
  \widehat{W}(e) = \mathbb{P}(\mathcal{C}(e_x) \neq \mathcal{C}(e_y))
\end{equation} 
where $\mathcal{C}(.)$ denotes the class to which the vertex belongs. Before stating the algorithm, we review the \emph{seeded watershed} algorithm \cite{IEEE:jounral/tpami/cousty09} in  algorithm \ref{alg:0}. Stated simply, seeded watershed labels all the vertices greedily using the Euclidean edge-weights.

\begin{algorithm}[H]
  \caption{Seeded Watershed}
  \label{alg:0}
\begin{algorithmic}[1]
  \renewcommand{\algorithmicrequire}{\textbf{Input:}}
  \renewcommand{\algorithmicensure}{\textbf{Output:}}
  \REQUIRE $V$ is the set of vertices, $E$ is the set of edges and $X$ denotes the HSI dataset. A subset of labelled points $V_k \subset V$.
  \ENSURE  Labels for each of the vertices $\widehat{\mathcal{C}}(.)$
  \STATE Initialize the labels $\widehat{\mathcal{C}}(v)$ for all $v \in V_k$, and $\widehat{\mathcal{C}}(v) = \textbf{NULL}$ for all $v \notin V_k$.
  \STATE Set $W(e_x, e_y) = \| X(e_x) - X(e_y) \|$, i.e initialize the edge weights using Euclidean distances for all edges $e = (e_x, e_y)$. 
  \STATE Sort the edges $E$ in increasing order w.r.t $W$.
  \STATE Initialize the union-find data structure \textsc{UF},
  \FOR {$e = (e_x,e_y)$ in sorted edge set $E$}
    \IF {both $e_x$ and $e_y$ are labelled}
      \STATE do nothing
    \ELSE
      \STATE \textsc{UF}.union$(e_x, e_y)$
      \STATE Assign same label for $e_x$ and $e_y$.
    \ENDIF
  \ENDFOR
  \STATE Label each vertex of the connected component using labels $V_k$. \label{line:labelling}
  \RETURN Labels of the vertices.
\end{algorithmic} 
\end{algorithm}

\emph{Stochastic Watershed} uses the seeded watershed algorithm using random subset of features and random subset of vertices as seeds. Algorithm \ref{alg:1} ensembles several stochastic watersheds. Each vertex within the sampled vertices is given a distinct label. These labels are propagated using the seeded watershed. This procedure is repeated several times to obtain ESW (Ensemble Stochastic Watershed) Edge-Weights.

\begin{figure}
  \centering
  \includegraphics[width=0.5\linewidth]{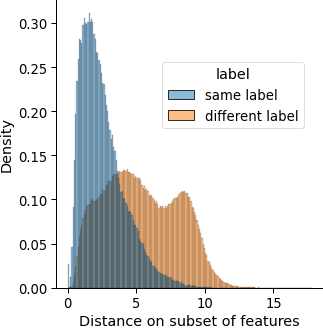}
  \caption{Distribution of edge-weights for same class and different class on Indianpines dataset. Here we plot this histogram of Euclidean distances on various subset of features between vertices of same/different class. Observe that the edge-weights for same class lie at the lower end, while edge-weights for different class are uniformly distributed.}
  \label{fig:2}
\end{figure}

\subsection*{Why does algorithm \ref{alg:1} work?}

The two main insights which lead to algorithm \ref{alg:1} are
\begin{enumerate}[label=(\roman*)]
  \item A commonly used statistical observation - averaging over samples from related distribution reduces the variance.
  \item Typically, one expects that feature distances and classes have two sided equivalence - That is, similar features imply same classes and same classes imply similar features. However, in case of HSI datasets, we have that on a subset of features
  \begin{eqnarray}
  \text{Similar features on a subset} &\Leftarrow \text{Same Class} \label{eq:same class imply sim feat} \\
\text{Similar features on a subset} &\not\Rightarrow \text{Same Class} \label{eq:sim feat not imply same class}
\end{eqnarray}
  This is verified in figure \ref{fig:2}. Observe that the edge-weights for same class lie at the lower end, while edge-weights for different class are uniformly distributed. Thus, \emph{one cannot simply consider an average of edge weights} for ensembles. Moreover, note that average over Euclidean distances using random subset of features result in Euclidean distances, that is
  \begin{equation}
    \frac{1}{N} \sum_{i=1}^{N} \|X_{s_i}(e_x) - X_{s_i}(e_y)\|^2 \approx \|X(e_x) - X(e_y)\|^2
  \end{equation}
where $e = (e_x, e_y)$ denotes a specific edge, $s_i$ denotes a random subset of features and $N$ denotes the size of the ensemble. 
\end{enumerate}

\begin{algorithm}[H]
  \caption{Computing ESW (Ensemble Stochastic Watershed) Edge-Weights}
  \label{alg:1}
\begin{algorithmic}[1]
  \renewcommand{\algorithmicrequire}{\textbf{Input:}}
  \renewcommand{\algorithmicensure}{\textbf{Output:}}
  \REQUIRE $V$ is the set of vertices, $E$ is the set of edges and $X$ denotes the HSI dataset. $N$ denotes the number of repetitions.
  \ENSURE  Estimated weights, $\widehat{W} : E \to \mathbb{R}^{+}$
  \STATE Initialize $\widehat{W}(e) = 0$ for all edges $e$. Let $c$ (initialized to $0$) denote the count.
  \REPEAT
    \STATE Select a subset of features of size $\kappa_f$, $S_{k}$. \label{alg:1 line:1} 
    \STATE Select a random sample from  the set of vertices of size $\kappa_v$, $V_{k}$. \label{alg:1 line:2}
    \STATE Assign distinct label (pseudo-labels) to each randomly selected vertex. There would be $\kappa_v$ distinct labels. \label{alg:1 line:3}
    \STATE Propagate the labels using \emph{seeded watershed} as described in algorithm \ref{alg:0}. Use the subset of sampled vertices $(V_{k})$ as seeds and the sampled subset of features $(S_{k})$ for computing edge-weights. Let $\widehat{\mathcal{C}}(.)$ denote the labelling obtained. \label{alg:1 line:4}
    \STATE Add $1$ to the estimated edge weight if the labels of its vertices are different. That is, 
      \begin{equation*}
        \widehat{W}(e_x,e_y) += I(\widehat{\mathcal{C}}(e_x) \neq \widehat{\mathcal{C}}(e_y))
      \end{equation*}
      where $I(.)$ denotes the indicator function. Also, add $1$ to the count $c$. \label{alg:1 line:5}
    \UNTIL{$c$ is less than $N$} \label{alg:1 line:6}
  \STATE Set $\widehat{W}(e) = \widehat{W}(e)/N$ to obtain an estimate between $[0,1]$.\label{alg:1 line:7}
\end{algorithmic} 
\end{algorithm}

In this article we consider an alternate approach to estimating edge-weights using ensembles of stochastic watershed, described in algorithm \ref{alg:1}. Here we consider the artificial labels constructed using the random samples and features (lines \ref{alg:1 line:2}, \ref{alg:1 line:3} and \ref{alg:1 line:4} of algorithm \ref{alg:1}). These artificial labels are then converted to edge-weights using the indicator function (line \ref{alg:1 line:5} in algorithm \ref{alg:1}). These edge-weights are then averaged to obtain the final estimate (line \ref{alg:1 line:7} in algorithm \ref{alg:1}).

Figure \ref{fig:1} visualizes the edge weights obtained using algorithm \ref{alg:1}. In what follows, we quantitatively evaluate the edge weights obtained using Random Walk (RW) and vanilla Graph Convolution Network (GCN). The code is available at \url{https://github.com/ac20/EnsembleEdgeWeightsHSI}

{\noindent
\textbf{Remark on Datasets : } We consider the following three HSI datasets for evaluation. All these datasets have been downloaded from \url{http://www.ehu.eus/ccwintco/index.php/Hyperspectral_Remote_Sensing_Scenes}
\begin{enumerate}
  \item \textbf{Indianpines :} This is  acquired by AVRIS spectograph, and has the size $145\times 145$. There are $200$ spectral bands. The groundtruth consists of $16$ different land cover classes. 
  \item \textbf{Salinas :} This is  acquired by AVRIS spectograph, and has the size  $512\times 217$. There are $204$ spectral bands. The groundtruth consists of $16$ different land cover classes.
  \item \textbf{Pavia Centre :} This is acquired by the ROSIS sensor and has the size $1096 \times 490$. There are $102$ spectral bands and groundtruth consists of $9$ different land cover classes.
\end{enumerate}
}

\section{Evaluating ESW Edge-Weights using Random Walk (Semi-Supervised)}
\label{sec:random walker}

\begin{figure*}
  \centering
  \subfloat[Indianpines]{\includegraphics[width=0.25\linewidth]{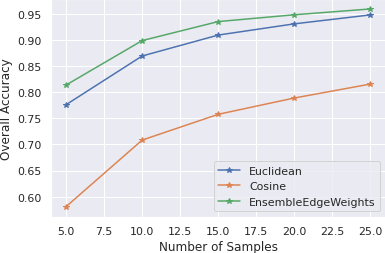}%
  \label{fig:3a}}
  \hfil
  \subfloat[Pavia Centre]{\includegraphics[width=0.25\linewidth]{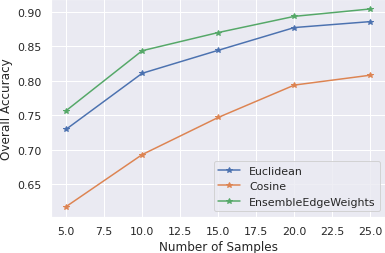}%
  \label{fig:3b}}
  \hfil
  \subfloat[Salinas GT]{\includegraphics[width=0.25\linewidth]{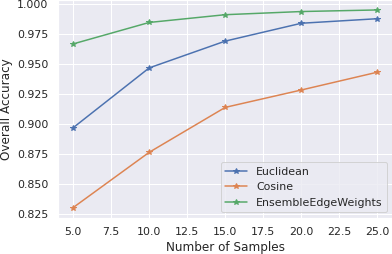}%
  \label{fig:3c}}

  \caption{Random Walk evaluation of ESW Edge-Weights. We compare the proposed edge-weights with the traditional edge-weights obtained by Euclidean distances \eqref{eq:euclidean sim} and cosine similarity \eqref{eq:cosine sim}. We evaluate using (a) Indianpines dataset. (b) Pavia Centre dataset and (c) Salinas dataset. Observe that in all cases ESW Edge-weights outperform the traditional measures.}
  \label{fig:3}
\end{figure*}

Random Walk (RW) is a classical tool for image segmentation \cite{IEEE:jounral/tpami/grady06}. It uses an edge-weighted graph and few labelled pixels to obtain the segmentation of the image. This is similar to the semi-supervised learning paradigm. In this section we compare our edge-weights - ESW Edge-Weights with the classic measures - Euclidean distance and cosine similarity. 

{\noindent
\textbf{Remark:} For sake of simplicity we describe the RW procedure for 2 labels - $\{0,1\}$. It extends to multi-label scenario naturally as described in \cite{IEEE:jounral/tpami/grady06}
}

To recap, let $G=(V, E, W)$ denote the edge-weighted graph. Construct the Laplacian matrix (called Random Walk Laplacian \cite{SPRINGER:journal/statcomp/Luxburg07}), $L$, where $L_{ij} = \sum_j w_{ij}$ if $i=j$ and $L_{ij} = -w_{ij}$ if $i \neq j$. Here $w_{ij}$ indicates the edge-weight between pixel $i$ and $j$. Let $S_0$ denote the set of vertex-indices labelled $0$, $S_1$ denote the set of vertex-indices labelled $1$, and $U$ denote the unlabelled vertex-indices. We then solve
\begin{equation}
  \begin{aligned}
    & \underset{x}{\text{minimize}}
    & &Tr(x^t L x) \\
    & \text{subject to}
    & & x[S_0] = 0 \text{ and } x[S_1] = 1 \\
    \end{aligned}
\end{equation}
Classical approaches to obtain edge-weights $w_{ij}$ are - (a) Using Euclidean distances
\begin{equation}
  w_{ij} = \exp(-\beta\|X(i) - X(j) \|))
  \label{eq:euclidean sim}
\end{equation}
where $X(i)$ indicates the feature vector of pixel $i$ and $\|X(i) - X(j) \|$ indicates the Euclidean distance, and (b) Cosine similarity 
\begin{equation}
  w_{ij} = \frac{X(i)^{t} X(j)}{\|X(i)\|\|X(j)\|}
  \label{eq:cosine sim}
\end{equation}

We compare the proposed ESW edge-weights obtained by algorithm \ref{alg:1} with the Euclidean edge-weights \eqref{eq:euclidean sim} and cosine edge-weights \eqref{eq:cosine sim}. Figure \ref{fig:3} shows the plots of overall accuracy versus number of samples used as seeds. Observe that in all these cases, the proposed edge-weights outperform the traditional methods. {(\textbf{Remark:} These results are obtained using the average score over 50 iterations.)}

\section{Evaluating ESW Edge-Weights using vanilla Graph Convolution Network (Unsupervised)}

While the previous section uses a semi-supervised technique for evaluating edge-weights, this section uses an unsupervised method - Graph Convolution Networks (GCN). We use the most basic version of GCN as described in \cite{ACM:conf/ijcai/Xiaotong19}. The approach described in \cite{ACM:conf/ijcai/Xiaotong19} does not use edge-weights. Here, we modify the approach to incorporate edge-weights.

Let $G = (V, E)$ denote the un-weighted graph. The normalized graph Laplacian $L^{norm}$ is defined as 
\begin{equation}
  L^{norm}_{ij} = \begin{cases}
    1 & \text{ if } i=j \\
    -\frac{w_{ij}}{\sqrt{\sum_j w_{ij}}\sqrt{\sum_i w_{ij}}} & \text{ if } i \neq j
  \end{cases}
  \label{eq:norm laplacian}
\end{equation}
One-step graph convolution is defined as
\begin{equation}
  X^{(k+1)} = (I - \frac{1}{2}L^{norm}) X^{(k)}
\end{equation}
Where $X^{(k)}$ denote the features after $k$ steps of convolution and $X^{(0)} = X$ (original data). These features are then used for spectral clustering \cite{SPRINGER:journal/statcomp/Luxburg07} to obtain class labels.

To understand the intuition behind the above formalism - Consider the unweighted graph where $w_{ij} = 1$ if there exists an edge between $(i,j)$ and $0$ otherwise.  In this case, one can show that the eigenvalues for the Laplacian in \eqref{eq:norm laplacian} belong to the interval $[0,2]$. Considering the spectral decomposition of $L^{norm}$, we have
\begin{equation}
  I - \frac{1}{2}L^{norm} = I - \frac{1}{2} U\Lambda U^{-} = U(I - \frac{1}{2}\Lambda)U^{-}
\end{equation}
Here $\Lambda$ is a diagonal matrix of eigenvalues of $L^{norm}$. Thus, the eigenvalues of $I - \frac{1}{2}L^{norm}$ are less than $1$. Thus, multiplication with $I - \frac{1}{2}L^{norm}$ will result in a \emph{low-pass} filter. (\textbf{Remark:} details can be found in \cite{ACM:conf/ijcai/Xiaotong19}).

We now modify the above approach for edge-weighted graph. For general edge-weights $w_{ij}$ eigenvalues of Laplacian  no longer belong to $[0,2]$, but belong to $[0, \lambda_{max}]$ where $\lambda_{max}$ denotes the maximum eigenvalue. Thus we define the one step graph convolution as 
\begin{equation}
  X^{(k+1)} = (I - \frac{1}{\lambda_{max}}L^{norm}) X^{(k)}
\end{equation}
These features are then used for spectral clustering to obtain class labels.

\begin{figure*}
  \centering
  \subfloat[Indianpines]{\includegraphics[width=0.25\linewidth]{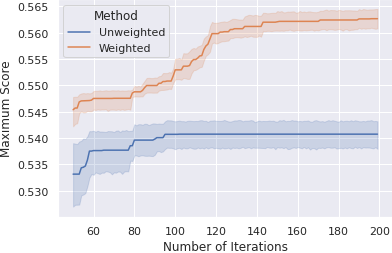}%
  \label{fig:4a}}
  \hfil
  \subfloat[Pavia Centre]{\includegraphics[width=0.25\linewidth]{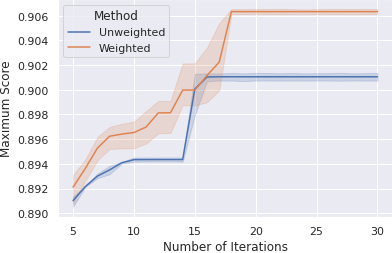}%
  \label{fig:4b}}
  \hfil
  \subfloat[Salinas GT]{\includegraphics[width=0.25\linewidth]{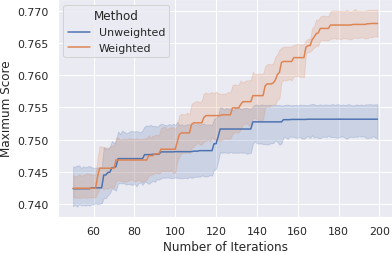}%
  \label{fig:4c}}

  \caption{GCN evaluation of ESW Edge-weights. We use the vanilla GCN \cite{ACM:conf/ijcai/Xiaotong19} modified to incorporate edge-weights. We plot the best score till iteration $k$. The number of iterations is shown on the x-axis and the max score is shown on the y-axis. Three datasets were used - (a) Indianpines (b) Pavia Centre and (d) Salinas. For Pavia Centre we stop after iteration $30$ since both methods have converged. Observe that the weighted GCN has significant improvement over unweighted GCN.}
  \label{fig:4}
\end{figure*}

{\noindent
  \textbf{Evaluation Procedure :} To evaluate the proposed ESW Edge-weights we perform GCN with and without weights for $200$ iterations. After each iteration, the spectral clustering is performed to get the clusters. These clusters are matched with the groundtruth clusters using Hungarian algorithm \cite{DOI:journal/misc/kuhn55}. Overall accuracy (OA) is measured as a percentage of samples correctly classified. OA is computed after every iteration and best score is recorded. 
}

The above procedure is repeated $10$ times and the average is computed for each iteration. This is plotted in figure \ref{fig:4}. We compare the proposed edge-weights with unweighted GCN. The proposed edge-weights provides significant improvement over the unweighted approach. Table \ref{table:1} reports the optimal results we obtained for each dataset. As a baseline we provide the results from \cite{IEEE:jounral/tgrs/Nan21} which is the recent state-of-the-art. Observe that proposed Weighted GCN outperforms the baseline method. 

\begin{table}[!t]
  \caption{\label{table:1} Overall Accuracy obtained by using the proposed edge-weights. Observe that using the proposed edge-weights gives significant improvements over unweighted approach. In all cases, the proposed method is comparable to thc current state-of-the-art in \cite{IEEE:jounral/tgrs/Nan21}.}
  \centering
  \begin{tabular}{lccc}\toprule
   Method & Indianpines & PaviaCentre & Salinas\\
  \midrule
  GCN (Unweighted) & 54.07 & 90.09& 75.31\\
  GCN (Weighted) & 56.26 & 90.62&76.80\\
  Baseline \cite{IEEE:jounral/tgrs/Nan21} & 53.09 & 87.72 &76.66\\
  \bottomrule
  \end{tabular}
  \end{table}

\section{Conclusion and Perspectives}

To summarize, in this article we propose a novel approach to estimate the edge-weights of the HSI Datasets. This approach exploits 2 fundamental insights - (a) Ensembles reduce the variance and (b) In hyperspectral datasets we have
\begin{equation}
  \begin{aligned}
    \text{Similar Features} &\Leftarrow \text{Same Class} \\
    \text{Similar Features} &\not\Rightarrow \text{Same Class}
  \end{aligned}
\end{equation}
These insights lead to algorithm \ref{alg:1}, where we instead average propagated pseudo-labels. To analyze the accuracy of the proposed edge-weights, we consider two evaluation approaches covering both semi-supervised and unsupervised scenarios. 

Firstly, using accuracies obtained via Random-Walk, we consider how well the proposed edge-weights compare with the traditional Euclidean distances and cosine similarities. We observe that the proposed edge-weights outperform the other measures at varying number of samples. Next unsupervised classification using vanilla GCN (as proposed in \cite{ACM:conf/ijcai/Xiaotong19}) is considered. Since, the approach in \cite{ACM:conf/ijcai/Xiaotong19} does not consider edge-weights, the method is suitably adapted to incorporate edge-weights. We show that proposed edge-weighted GCN outperforms the unweighted GCN and recent state-of-the-art \cite{IEEE:jounral/tgrs/Nan21}.

The main outcome of the above article is the emphasis on using edge-weights. While several works use graphs and related approaches, not much work was done to incorporate edge-weights. In this article we estimated the edge-weights explicitly using ensembles of stochastic watershed. Incorporating these edge-weights improved the results of the existing approaches such as GCN. This is mainly due to the fact that edge-weights have the ability to capture not just second order relations (edges) but also higher order relations (hyperedges). Adapting other graph based approaches to incorporate edge-weights is considered for future work.

% \appendices
% \section{Details about Morphological Distances}

% % you can choose not to have a title for an appendix
% % if you want by leaving the argument blank
% \section{}

\section*{Acknowledgment}

AC and SD would like to thank APPCAIR and BITS Pilani K K Birla Goa Campus. The work of AC was supported by BITS-Pilani, K. K. Birla Goa Campus, under Grant BPGC/RIG/2021-22/09-2021/02. SD would like to acknowledge the funding received from BPGC/RIG/2020-21/11-2020/01 (Research Initiation Grant), GOA/ACG/2021-22/Nov/05 (Additional Competitive Grant) both provided by BITS-Pilani K K Birla Goa Campus.

% \ifCLASSOPTIONcaptionsoff
%   \newpage
% \fi

\bibliographystyle{IEEEtran}
\bibliography{references}

\begin{thebibliography}{10}
\providecommand{\url}[1]{#1}
\csname url@samestyle\endcsname
\providecommand{\newblock}{\relax}
\providecommand{\bibinfo}[2]{#2}
\providecommand{\BIBentrySTDinterwordspacing}{\spaceskip=0pt\relax}
\providecommand{\BIBentryALTinterwordstretchfactor}{4}
\providecommand{\BIBentryALTinterwordspacing}{\spaceskip=\fontdimen2\font plus
\BIBentryALTinterwordstretchfactor\fontdimen3\font minus
  \fontdimen4\font\relax}
\providecommand{\BIBforeignlanguage}[2]{{%
\expandafter\ifx\csname l@#1\endcsname\relax
\typeout{** WARNING: IEEEtran.bst: No hyphenation pattern has been}%
\typeout{** loaded for the language `#1'. Using the pattern for}%
\typeout{** the default language instead.}%
\else
\language=\csname l@#1\endcsname
\fi
#2}}
\providecommand{\BIBdecl}{\relax}
\BIBdecl

\bibitem{IEEE:jounral/tgrs/Nan21}
N.~Huang, L.~Xiao, J.~Liu, and J.~Chanussot, ``Graph convolutional sparse
  subspace coclustering with nonnegative orthogonal factorization for large
  hyperspectral images,'' \emph{IEEE Transactions on Geoscience and Remote
  Sensing}, pp. 1--16, 2021.

\bibitem{IEEE:jounral/tgrs/Yaoming21}
Y.~Cai, Z.~Zhang, Z.~Cai, X.~Liu, X.~Jiang, and Q.~Yan, ``Graph convolutional
  subspace clustering: A robust subspace clustering framework for hyperspectral
  image,'' \emph{IEEE Transactions on Geoscience and Remote Sensing}, vol.~59,
  no.~5, pp. 4191--4202, 2021.

\bibitem{SPRINGER:journal/Multimedia/Patel22}
\BIBentryALTinterwordspacing
H.~Patel and K.~P. Upla, ``A shallow network for hyperspectral image
  classification using an autoencoder with convolutional neural network,''
  \emph{Multimedia Tools and Applications}, vol.~81, no.~1, pp. 695--714, 2022.
  [Online]. Available: \url{https://doi.org/10.1007/s11042-021-11422-w}
\BIBentrySTDinterwordspacing

\bibitem{IEEE:journal/grsm/Ghamisi17}
P.~Ghamisi, J.~Plaza, Y.~Chen, J.~Li, and A.~J. Plaza, ``Advanced spectral
  classifiers for hyperspectral images: A review,'' \emph{IEEE Geoscience and
  Remote Sensing Magazine}, vol.~5, no.~1, pp. 8--32, 2017.

\bibitem{IEEE:journal/tgrs/Kruse03}
F.~Kruse, J.~Boardman, and J.~Huntington, ``Comparison of airborne
  hyperspectral data and eo-1 hyperion for mineral mapping,'' \emph{IEEE
  Transactions on Geoscience and Remote Sensing}, vol.~41, no.~6, pp.
  1388--1400, 2003.

\bibitem{IEEE:journal/tgrs/Shimoni19}
M.~Shimoni, R.~Haelterman, and C.~Perneel, ``Hypersectral imaging for military
  and security applications: Combining myriad processing and sensing
  techniques,'' \emph{IEEE Geoscience and Remote Sensing Magazine}, vol.~7,
  no.~2, pp. 101--117, 2019.

\bibitem{IEEE:journal/grsl/Lixia14}
L.~Yang, S.~Yang, P.~Jin, and R.~Zhang, ``Semi-supervised hyperspectral image
  classification using spatio-spectral laplacian support vector machine,''
  \emph{IEEE Geoscience and Remote Sensing Letters}, vol.~11, no.~3, pp.
  651--655, 2014.

\bibitem{IEEE:journal/jstars/Chen19}
C.~Liu, J.~Li, and L.~He, ``Superpixel-based semisupervised active learning for
  hyperspectral image classification,'' \emph{IEEE Journal of Selected Topics
  in Applied Earth Observations and Remote Sensing}, vol.~12, no.~1, pp.
  357--370, 2019.

\bibitem{IEEE:journal/tip/Hao18}
H.~Wu and S.~Prasad, ``Semi-supervised deep learning using pseudo labels for
  hyperspectral image classification,'' \emph{IEEE Transactions on Image
  Processing}, vol.~27, no.~3, pp. 1259--1270, 2018.

\bibitem{IEEE:journal/grsm/Han21}
H.~Zhai, H.~Zhang, P.~LI, and L.~Zhang, ``Hyperspectral image clustering:
  Current achievements and future lines,'' \emph{IEEE Geoscience and Remote
  Sensing Magazine}, pp. 0--0, 2021.

\bibitem{IEEE:journal/tgrs/Valls07}
G.~Camps-Valls, T.~V. Bandos~Marsheva, and D.~Zhou, ``Semi-supervised
  graph-based hyperspectral image classification,'' \emph{IEEE Transactions on
  Geoscience and Remote Sensing}, vol.~45, no.~10, pp. 3044--3054, 2007.

\bibitem{IEEE:journal/tgrs/Yao21}
Y.~Ding, X.~Zhao, Z.~Zhang, W.~Cai, N.~Yang, and Y.~Zhan, ``Semi-supervised
  locality preserving dense graph neural network with arma filters and
  context-aware learning for hyperspectral image classification,'' \emph{IEEE
  Transactions on Geoscience and Remote Sensing}, pp. 1--12, 2021.

\bibitem{IEEE:journal/jstars/Haoyu21}
H.~Wang, Y.~Cheng, C.~L.~P. Chen, and X.~Wang, ``Semisupervised classification
  of hyperspectral image based on graph convolutional broad network,''
  \emph{IEEE Journal of Selected Topics in Applied Earth Observations and
  Remote Sensing}, vol.~14, pp. 2995--3005, 2021.

\bibitem{ELSIVIER:journal/pr/Sellami22}
\BIBentryALTinterwordspacing
A.~Sellami and S.~Tabbone, ``Deep neural networks-based relevant latent
  representation learning for hyperspectral image classification,''
  \emph{Pattern Recognition}, vol. 121, p. 108224, 2022. [Online]. Available:
  \url{https://www.sciencedirect.com/science/article/pii/S0031320321004052}
\BIBentrySTDinterwordspacing

\bibitem{SPRINGER:journal/statcomp/Luxburg07}
\BIBentryALTinterwordspacing
U.~von Luxburg, ``A tutorial on spectral clustering,'' \emph{Statistics and
  Computing}, vol.~17, no.~4, pp. 395--416, 2007. [Online]. Available:
  \url{https://doi.org/10.1007/s11222-007-9033-z}
\BIBentrySTDinterwordspacing

\bibitem{IEEE:journal/tgrs/Luo16}
F.~Luo, H.~Huang, Z.~Ma, and J.~Liu, ``Semisupervised sparse manifold
  discriminative analysis for feature extraction of hyperspectral images,''
  \emph{IEEE Transactions on Geoscience and Remote Sensing}, vol.~54, no.~10,
  pp. 6197--6211, 2016.

\bibitem{IEEE:journal/tgrs/Luo22}
F.~Luo, Z.~Zou, J.~Liu, and Z.~Lin, ``Dimensionality reduction and
  classification of hyperspectral image via multistructure unified
  discriminative embedding,'' \emph{IEEE Transactions on Geoscience and Remote
  Sensing}, vol.~60, pp. 1--16, 2022.

\bibitem{ACM:conf/ijcai/Xiaotong19}
X.~Zhang, H.~Liu, Q.~Li, and X.-M. Wu, ``Attributed graph clustering via
  adaptive graph convolution,'' in \emph{Proceedings of the 28th International
  Joint Conference on Artificial Intelligence}, ser. IJCAI'19.\hskip 1em plus
  0.5em minus 0.4em\relax AAAI Press, 2019, p. 4327–4333.

\bibitem{IEEE:jounral/tpami/cousty09}
J.~Cousty, G.~Bertrand, L.~Najman, and M.~Couprie, ``Watershed cuts: Minimum
  spanning forests and the drop of water principle,'' \emph{IEEE Transactions
  on Pattern Analysis and Machine Intelligence}, vol.~31, no.~8, pp.
  1362--1374, 2009.

\bibitem{SCIENCEDIRECT:journal/cvgip/kovalevsky89}
\BIBentryALTinterwordspacing
V.~Kovalevsky, ``Finite topology as applied to image analysis,'' \emph{Computer
  Vision, Graphics, and Image Processing}, vol.~46, no.~2, pp. 141--161, 1989.
  [Online]. Available:
  \url{https://www.sciencedirect.com/science/article/pii/0734189X89901655}
\BIBentrySTDinterwordspacing

\bibitem{IEEE:jounral/tpami/grady06}
L.~Grady, ``Random walks for image segmentation,'' \emph{IEEE Transactions on
  Pattern Analysis and Machine Intelligence}, vol.~28, no.~11, pp. 1768--1783,
  2006.

\bibitem{DOI:journal/misc/kuhn55}
\BIBentryALTinterwordspacing
H.~W. Kuhn, ``The hungarian method for the assignment problem,'' \emph{Naval
  Research Logistics Quarterly}, vol.~2, no. 1–2, p. 83–97, Mar 1955.
  [Online]. Available: \url{http://dx.doi.org/10.1002/nav.3800020109}
\BIBentrySTDinterwordspacing

\end{thebibliography}

% \end{IEEEbiography}


% % insert where needed to balance the two columns on the last page with
% %\newpage



%\vfill


\end{document}